\RequirePackage{fix-cm}
\documentclass[smallcondensed]{svjour3}     
\smartqed  
\usepackage[numbers]{natbib}
\usepackage{float}
\usepackage{graphicx}
\usepackage{multirow}
\usepackage{url}
\usepackage{subfigure}
\usepackage{soul}
\usepackage{xcolor}
\journalname{}
\begin{document}

\title{Symbolic regression by uniform random global search}


\author{Sohrab Towfighi}


\institute{1 King's College Circle, Toronto, ON M5S 1A8 \\
              Faculty of Medicine,               University of Toronto\\
              \email{sohrab.towfighi@mail.utoronto.ca}             \\
              ORCID: 0000-0002-3050-8943
}

\date{Received: date / Accepted: date}

\maketitle

\begin{abstract} 
Symbolic regression (SR) is a data analysis problem where we search for the mathematical expression that best fits a numerical dataset. It is a global optimization problem. The most popular approach to SR is by genetic programming (SRGP). It is a common paradigm to compare an algorithm's performance to that of random search, but the data comparing SRGP to random search is lacking. We describe a novel algorithm for SR, namely SR by uniform random global search (SRURGS), also known as pure random search. We conduct experiments comparing SRURGS with SRGP using 100 randomly generated equations. Our results suggest that a SRGP is faster than SRURGS in producing equations with good $R^2$ for simple problems. However, our experiments suggest that SRURGS is more robust than SRGP, able to produce good output in more challenging problems. As SRURGS is arguably the simplest global search algorithm, we believe it should serve as a control algorithm against which other symbolic regression algorithms are compared. SRURGS has only one tuning parameter, and is conceptually very simple, making it a useful tool in solving SR problems. The method produces random equations, which is useful for the generation of symbolic regression benchmark problems. We have released well documented and open-source python code, currently under formal peer-review, so that interested researchers can deploy the tool in practice.


\paragraph{}
\textbf{Article Highlights}
\begin{itemize}
    \item an algorithm for finding the equation of best fit is presented
    \item the algorithm can generate benchmark problems
    \item experiments show the algorithm is competitive with genetic programming
\end{itemize}

\end{abstract}

\vfil
\section{Introduction}

In research, workers produce numerical data that describes natural phenomena. Researchers often needs to determine the most appropriate equation to describe the data set. Symbolic regression is an analysis which searches for the ideal equation to represent a given data set. The problem appears to be NP hard, but we find no formal proof. The term symbolic regression seems to have been coined in the seminal paper introducing genetic programming \cite{Koza1994}, which partly explains why genetic programming is the du jour methodology used in solving symbolic regression problems. Genetic programming has its limitations, and a few researchers have explored whether there are problems for which genetic programming performs worse than does random search. We refer to evolutionary computing for solving symbolic regression problems as symbolic regression by genetic programming (SRGP). SRGP has been extensively studied, but has drawbacks: it is slow, and it has many hyperparameters which are of unknown importance to lay users. This work considers a new approach to symbolic regression, namely symbolic
5 regression by uniform random global search. In broad strokes, the algorithm works by asking the user to specify the most complex binary tree they want to be considered in the search, then the algorithm counts all the possible equations using the simplest binary tree up to the most complex one permitted, then it randomly selects from these equations giving each one the same probability of being selected.

\subsection{Genetic Algorithms vs. Random Search}

We have found one study comparing random search to genetic programming for symbolic regression \cite{Amaral:1997}, but they do not give detail regarding the form of random search they use nor do they conduct extensive experiments. These workers found that the data structure being used to represent equations plays a pivotal role in the performance of genetic algorithms. They state that the binary string representation can be subject to a great deal of deception, which describes when two well performing individuals reproduce to give an ill performing offspring. Ultimately, they argue that a well tuned genetic algorithm can outperform random search, but do not provide strong evidence for this claim. As readers, we asked ourselves how a practitioner would go about appropriately tuning their genetic algorithm prior to executing their symbolic regression search. There is a chicken and egg problem here whereby you need to tune the model before running it against your dataset, but tuning implies running the algorithm for multiple configurations against the dataset.

There is evidence that random search can perform similarly to or better than genetic algorithms in various problem domains \cite{Kozak:2014,Shamshiri:2018, Jalote:2014, randoGenFin, Amaral:1997}. For the problem of performing stratified random sampling in statistics, it was found that a genetic algorithm approach performed similarly to or worse than random search \cite{Kozak:2014}. Surprisingly, the author still suggested that genetic algorithms are promising algorithms for this purpose. In software development, the production of a test suite is important in maintaining quality. In one study, several algorithms including an evolutionary algorithm and variants of random search were compared. The algorithms were comparable with respect to the percent of branch coverage that their test suites amassed \cite{Shamshiri:2018}. The evolutionary algorithm covered 69\%, a random search with enhanced seeding 69\%, and pure random search 65\%. Another project considered automated approaches to creating software patches \cite{Jalote:2014} and found that random search outperformed genetic programming in 23 of 24 cases. Their random search approach generated more effective patches. Appropriately, they suggest that optimization algorithms be consistently compared with random search to assess performance. In computational finance, algorithmic approaches can be used to determine the appropriate trading strategies. One research group describes studies on the performance of genetic programming in this field as typically inconclusive, but with the studies always arguing for a need for further study \cite{randoGenFin}. This group performed a study comparing genetic programming with equal intensity random search, meaning that the random search considers the same number of equations as does the genetic programming approach. They do not give great detail in describing their results or their random search methodology, but they do state that in cases where it is plausible that the training data has learn-able patterns, both random search and genetic programming outperform lottery trading.

There is a very large study investigating the parameter space of genetic algorithms and the effect of hyperparameter selection on algorithm performance \cite{Sipper2018}. These workers attempt to answer the question of whether the evolutionary component of SRGP, its crossover and mutation operations, confers any advantage over random search. To do so, they took their well performing parameter sets and for each of them, ran an equal intensity random search to see whether random search would be able to solve the benchmark problems they were considering. The number of models they allowed random search to consider equalled the size of the genetic algorithm's population times the number of generations. The authors assert that they examined 47,028,011 randomly generated models and none of the randomly generated models passed their success criteria. The authors state they used the evolutionary algorithm's built in method for initializing its population as the means of generating these randomly generated functions. We contend that these authors inadvertently limited the search space of the random search and so doomed the random search to failure. The authors used the Distributed Evolutionary Algorithms in Python's (DEAP \cite{FortinFelix-Antoine2012}) documentation's tutorials as a basis for their experiments, and a review of that code shows that the individuals randomly generated are limited to a maximum tree height of 2. We asked one of the authors if they indeed used that example from DEAP, and they were kind in their correspondence, describing how they implemented their search. We reproduced it and examined the generated models, which were too simple to capture the behaviour of the quartic polynomial, one of the problems they were considering in their analysis. Therefore, we do not believe this study provides evidence to suggest that random search performs worse than does genetic programming. 

Though publication bias is a well known phenomenon in biomedical literature, it has been found to be a problem in computing literature as well \cite{pubBiasComp, Nissen:2016}. Performing worse than random search could be considered a negative result, so workers in the field may feel pressure to avoid publishing experiments where random search outperforms their algorithm. There is tremendous pressure to publish positive findings, which leads researchers to mine their data for statistical significance. Therefore, we believe that the performance of SRGP versus SRURGS deserves consideration.

\subsection{Types of random search}
There are different types of random search. There is random local search, where a position in the global domain is randomly selected, then nearby solutions are randomly selected for consideration. There is also pure random search, which we use, that randomly selects a selection from the solution space without any regard to the previously considered solutions. Our approach has no dependence on the choice of initial conditions and cannot get stuck at local minima. Our approach has also been called pure random search \cite{Zabinsky:2011}. It has been proven that uniform random global search will converge on the global minimum solution as the number of function evaluations tends to infinity \cite{Zabinsky2003, randomConvergence}. This is reasonable, because uniform random global search, if run indefinitely, reduces to an exhaustive search. To our knowledge, there is no literature on symbolic regression by uniform random global search (SRURGS).

\subsection{No Free Lunch Theorems}
There has been a great deal of literature considering the No Free Lunch Theorems (NFLT) \cite{Woodward2003}. The NFLT essentially state that the performance of all general purpose optimization algorithms, when averaged over all the possible problems the solver could encounter, will be equivalent \cite{NFLT}. The NFLT argue that algorithms may perform better on a particular subtype of problem, but will necessarily perform worse on others. The NFLT have some conditions on which their validity is contingent, including that the algorithms do not revisit candidate solutions. There has been some work which proves there are specific types of problems for which NFLT do not hold \cite{NFLT_nothold, Coevolutionary_FLT}, but these seem to be the exception as opposed to the rule. 

\subsection{Alternative algorithms for symbolic regression}
In examining the literature, we have come across a variety of novel algorithms for SR in addition to the typical SRGP approach. There are also a wide variety of genetic programming inspired algorithms like Grammatical Evolution, Cartesian GP and Linear GP which we will not explore in this analysis. An approach to SR is Fast Function Extraction (FFX), a method that uses pathwise regularized learning to prune the search space \cite{FFx}. This approach is described as fast, scalable, and deterministic. FFX outputs a generalized linear model of particular pre-specified basis functions, incrementing the size of the model until the user specified maximum complexity is reached. FFX is able to compute very quickly, but it is unable to capture complicated nonlinear relationships entirely. FFX is available as an open source package. Another deterministic approach to SR, called Prioritized Grammar Enumeration (PGE) \cite{PGE_Worm}, maintained the general form of an SRGP algorithm, but replaced its random number generation and genetic operations with grammar enumeration rules and a pareto priority queue, which guide the search. PGE works by starting the search with basis functions, then growing the expression using its grammar enumeration techniques, which permits more complex functions than does FFX. The author describes a test suite in which PGE performs very well relative to other competing algorithms. This package was released as an open source library, but at present it appears the code has been left in a non-functioning state. Another approach to SR, based on dynamic programming, is the Interaction-Transformation Symbolic Regression algorithm (ITSR) \cite{ITSR}, which uses a custom data structure along with a greedy tree heuristic in order to bound the search space and determine the ideal equation from this smaller search space. ITSR is available as a functioning open source library. Another novel approach to SR is Evolutionary Feature Synthesis (EFS) \cite{Arnaldo}, which presupposes a linear functional form, but uses evolutionary computing to perform feature engineering, manipulating the independent variables into a form such that the encompassing linear regression can approximate nonlinear equations. More recently, workers used an advanced commercial mixed-integer nonlinear programming (MINLP) generalized solver against the symbolic regression problem with very promising results \cite{Cozad2018}. They argue that the use of MINLP offers guaranteed deterministic global optimal solutions to within a pre-specified tolerance of error. This approach is quite complex to deploy. The authors state that "the resulting symbolic regression model... has 70 continuous variables, 33 binary variables, and 1378 constraints." The authors use commercial software to solve the MINLP problem and provide no open-source code. In addition, though deterministic, there is no guarantee offered with regards to how long the code will need to run to reach its conclusion. Practically speaking, it may be technically challenging and cost prohibitive to deploy this approach for many SR practitioners.

\begin{figure*}[h]
    \centering
    \includegraphics[scale=0.3]{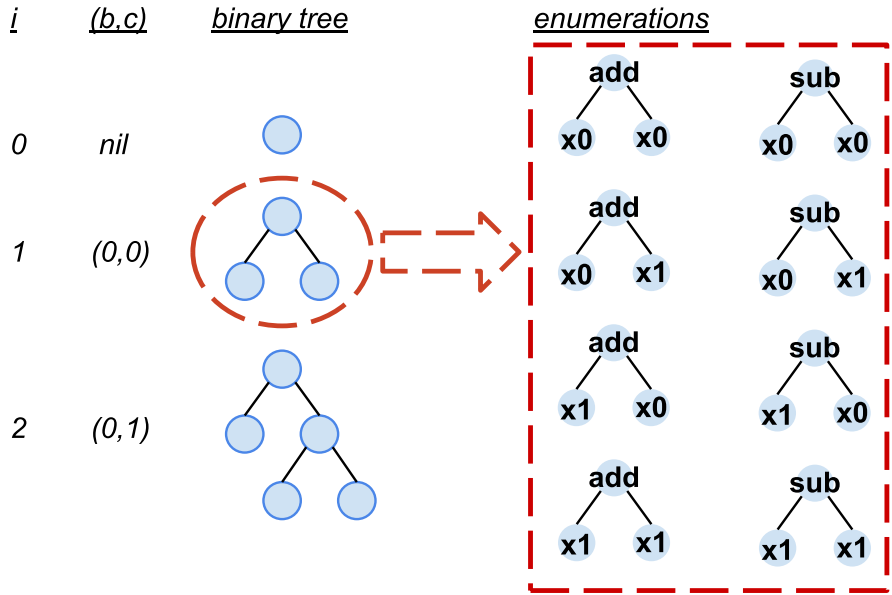}
    \caption{The methodology used to densely enumerate the symbolic regression problem space first specifies the shape of binary tree then specifies the particular configuration of the functions and terminals}
    \label{fig:method}
\end{figure*}

\subsection{Benchmarks in symbolic regression}

There has been recent work in the genetic programming literature on the quality and type of benchmark problems typically deployed in the field \cite{GPNeedsBetterBenchmarks, White2013}. They find that symbolic regression problems are the most commonly used, but they highlight areas where there could be improvement in the benchmarks used. They discuss good qualities in benchmarks: they are tunable in difficulty, fast to perform, varied in character, representation independent, accommodating to implementers, and relevant, among other qualities. They also discuss the need to avoid biasing algorithms towards a particular class of problems by "teaching to the test". They also argue against the use of metrics assessing how frequently the absolutely correct equation is reached, as they argue this promotes the use of toy problems. They argue that a better metric is to run the algorithm for a pre-specified number of iterations then measure the fitness of the best generated candidate solution and repeat this multiple times and take its average to calculate the best average fitness \cite{GPNeedsBetterBenchmarks, White2013}. Though SRGP is able to find good solutions, it has been found to fail to reach the exact solution in rather simple test problems \cite{Korns2011}. We think the random equation generator deployed in SRURGS makes for an ideal benchmark problem generator. We are able to tune the size of the produced equations. SRURGS' produced equations are varied and unbiased. Since SRURGS outputs equations as a string of python code, any symbolic regression platform should be able to use the benchmarks. 

\subsection{Symbolic regression search space}
The symbolic regression search space is discrete and discontinuous. Moreover, the search space has a large number of duplicate solutions. There is recent work which performed a cluster analysis of a univariate symbolic regression problem from a genetic programming context and found that the clustering of solutions based on genotypic similarity was poor, while the clustering of solutions based on phenotypic similarity was excellent, suggesting that the genotype landscape does not predict solution quality \cite{Kronberger2019}. In evolutionary algorithms, the genotype denotes the representation of the candidate solution in its data structure form, whereas the phenotype denotes the candidate solution once it has been evaluated. This result demonstrates that functions which appear similar in their data structure representation may ultimately produce vastly different outputs. A concept that is applicable is that of locality, which describes the extent to which candidate solutions are similar to their neighbouring candidate solutions within the search space and whether the fitness landscape directs the navigator towards its local trough. It has been found that for evolutionary algorithms to operate efficiently, the search space needs to have high locality \cite{Rothlauf:2006, GalvanLopez:2010}. 

\begin{figure}[h]
    \centering
    \includegraphics[scale=1.]{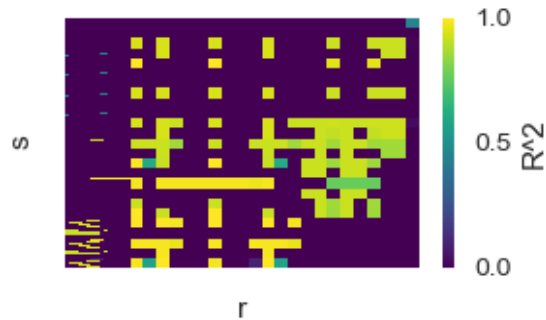}
    \caption{Heatmap showing coefficient of determination ($R^2$) of solutions in the classic quartic polynomial problem}
    \label{fig:quartic_polynom_domain}
\end{figure}

\subsection{Proposed experiments}
We begin with some computations that explore the solution space of the quartic polynomial problem, a toy problem that is commonly encountered in the symbolic regression literature. For our main analysis, we generate 100 randomly generated benchmark problems to evaluate SRURGS. Each benchmark problems is solved 10 times by SRGP and 10 times by SRURGS and comparison is made with respect to their performances. We do multiple runs for each algorithm because they are stochastic in nature.

We have discussed the need to evaluate the hegemony that genetic programming has maintained over the solution of SR problems. We are proposing, in concordance with the NFLT, that the SR analyst consider the use of SRURGS when facing their data analysis problems. SRURGS should approach convergence for all cases given sufficient computing time, and we will show that it generates good outputs in a reasonable number of iterations. In this work, we review the combinatorial nature of the solution space for SR problems, present an elegant and efficient method of enumerating the set of possible equations, leverage the machinery of this enumeration technique to perform uniform random global search from this set of possible equations, and deploy SRURGS against a database of 100 SR problems that we randomly generate. Our well documented code is publicly available under the GPL 3.0 license for interested researchers. The code is currently undergoing a formal peer review. \footnote{The source code for the algorithm along with documentation can be found at\\ \url{https://github.com/pySRURGS/pySRURGS}.}

\section{Methods}

We begin by describing the methodology we use to enumerate the SR problem space. We then move onto describing how we generate random equations.  We use this machinery in order to randomly generate a new benchmark problem set for use in comparing different SR algorithms. Some commonly used benchmark problem sets are \cite{Korns2011, NguyenSR}. We believe using these hand made benchmarks "teaches to the test", promoting the development of algorithms suited for the properties of these equations. This avoids the main challenge, which is to test that a SR algorithm is robust enough to perform well on any problem. Our random equation generation technique allows us to generate problems which are highly varied and without bias.

The data structure we use to represent mathematical equations is a binary tree. Each leaf of the tree corresponds to a terminal, which represents an independent variable in our dataset. The other nodes correspond to functions like addition, subtraction, multiplication, division, and power. These functions have an arity, which refers to the number of inputs the function takes. The addition function would have an arity of two, whereas trigonometric functions have an arity of one. We begin by describing how to enumerate the case where we only permit functions of arity two, but then move onto describing the more complex case including functions of arity one. Our code is able to handle both scenarios. We show equations for the enumeration of the problem space when functions of arity one and two are permitted. It is simple to extend this to the case where only functions of arity two are permitted so this is not shown.

\subsection{Enumerating binary trees}

The following maps non-negative integers to unique full binary trees \cite{Tychonievich2013}.

\begin{enumerate}
 \item If $i$ is zero then return a single terminal, else continue.
 \item If $i$ is one then return a non-terminal with two terminals as children, else continue.
 \item Convert $i-1$ to binary and call this binary number $a$.
 \item De-interleave $a$ to its odd and even bits.
 \item Convert the binary de-interleaved values to decimal. Refer to the odd bits value, now in decimal, as $b$, and refer to the even bits value, now in decimal, as $c$.
 \item Return a binary tree for its left child the tree corresponding to $i$ equals $b$ and for its right child the tree corresponding to $i$ equals $c$.
\end{enumerate}

This recursive relationship is valuable in minimizing the computational work associated with generating complex trees. We find that $b$ and $c$ are much smaller than $i$, so problem is reduced to two smaller problems. It is possible to save to permanent memory the shape of previously mapped values of i, so if the cases of $i$ equals $b$ and $i$ equals $c$ have already calculated, then the present computation is a mere matter of joining the two saved results. 

\subsection{Including functions of arity one}
The following is a modification of the previously discussed methodology to permit the case of functions of arity 1. This permits functions like tan, sin, or exp. We represent functions of arity one as a node with one valid child and a second child that is empty. For a given node in a tree, the case where the left child is empty is, practically speaking, identical to the case where the right child is empty. Therefore, our enumeration is designed to avoid the case where the left child is empty so as to prevent duplicating the problem space. The algorithm for this procedure is similar to the work of \cite{Tychonievich2013} and is presented below.

\begin{enumerate}
    \item If $i$ is zero then return a single terminal, else continue.
    \item If $i$ is one then return a non-terminal with two terminals as children, else continue.
    \item If $i$ is two then return a non-terminal with one terminal as its left child and None as its right child, else continue.
    \item Convert $i$ to binary and call this binary number $a$.
    \item De-interleave $a$ to its odd and even bits. 
    \item Convert the binary de-interleaved values to decimal. Refer to the odd bits value, now in decimal, as $b$, and refer to the even bits value, now in decimal, as $c$. 
    \item Return a binary tree for its left child the tree corresponding to $i$ equals $b$ and for its right child the tree corresponding to $i$ equals $c$.
\end{enumerate}

\begin{figure}[h]
    \centering
    \includegraphics[scale=1.]{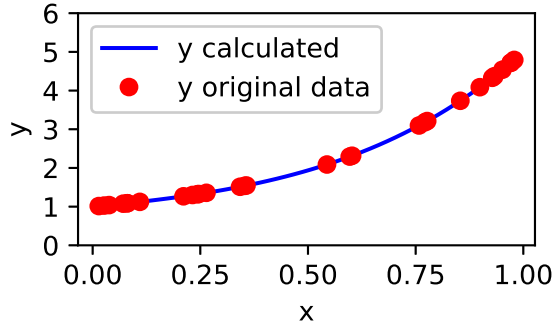}
    \caption{A plot demonstrating the goodness of fit for a SRURGS produced model against the quartic polynomial problem}
    \label{fig:quartic_polynom_result}
\end{figure}

\subsection{Enumerating the problem space}
Each binary tree $i$ has $l_i$ nodes in which to place functions of arity one. There are $f$ functions of arity one from which to choose. The number of possible arrangements of functions of arity one in the $i^{th}$ tree is $G_i$.

\begin{equation}\label{Eq_Gi}
    G_i = f^{l_i}
\end{equation}

Each binary tree $i$ has $k_i$ nodes in which to place functions of arity two. There are $n$ functions from which to choose. The number of possible arrangements of functions in the $i^{th}$ tree is $A_i$.

\begin{equation}\label{Eq_Ai}
    A_i = n^{k_i}
\end{equation}

Similarly, each binary tree has $j_i$ leaves in which to place terminals. There are $m$ terminals from which to choose. The number of possible arrangements of terminals, in the $i^{th}$ tree is $B_i$.

\begin{equation}\label{Eq_Bi}
    B_i = m^{j_i}
\end{equation}

To calculate the number of possible equations for the first $N$ binary trees, take the product of these two expressions and take their sum as $i$ iterates from zero through to $N - 1$.

\begin{equation}\label{Eq_M}
    M = \sum_{n=0}^{N-1}  f^{l_i} n^{k_i} m^{j_i}
\end{equation}
The number of possible equations for different use cases can be readily computed once a method is presented which solves for  $l_i$, $k_i$ and $j_i$. Of course, you could evaluate the shape of the $i^{th}$ tree and simply count the $l_i$, $k_i$ and $j_i$, but that would be inelegant. Leveraging the same recursive relationship found in the work of \cite{Tychonievich2013}, it is possible to calculate  $l_i$, $k_i$ and $j_i$

If we return to the method of \cite{Tychonievich2013} and examine how $l_i$ changes with $i$, while tracking $b$ and $c$, we find the following: the number of functions of arity one in a tree is equal to the sum of the number of functions of arity one in the root\textsc{\char13}s children, except for the case of $i$ equals 0, which represents a single terminal. It is possible to solve for $k_i$ and $j_i$ similarly. The number of functions of arity two in a tree is equal to the sum of the number of functions of arity two in the root\textsc{\char13}s children plus one, except for the case of $i$ equals zero, which represents a single terminal. The number of terminals in a tree is equal to the sum of the number of terminals in the root\textsc{\char13}s children, except for the case of $i$ equals zero. Now, we have a recursive method to solve for $l_i$, $j_i$ and $k_i$ and we are able to enumerate the number of possible equations.

\subsection{Generating random equations}
A random equation is specified when the tree is defined, the specific arrangement of the functions of arity one is defined, the specific arrangement of functions of arity two is defined, and the specific arrangement of the terminals is defined. An outline of the enumeration scheme is shown in Fig \ref{fig:method}. A tree is defined by the whole number $i$, the number of possible arrangements of functions of arity one $G_i$, the number of possible arrangements of functions of arity two $A_i$ and the number of possible arrangements of the terminals $B_i$ (see equations \ref{Eq_Gi},  \ref{Eq_Ai} and \ref{Eq_Bi}). If a value for $i$ is specified, a value within $[0, G_i-1]$ called $q$ can be randomly selected. Similarly, a value within $[0, A_i-1]$ called $r$ can be randomly selected, and a value within $[0, B_i-1]$ called $s$ can be randomly selected. These map to a particular equation configuration, leading to a fully specified random equation. 

There are some technical challenges associated with finding the $q^{th}$ configuration of the unit arity function arrangements, the $r^{th}$ configuration of the two arity function arrangements and the $s^{th}$ configuration of the terminal arrangements. In the Python programming language, the implementation of the itertools.product function returns a generator that allows you to iterate through all the possible arrangements, but if the number of arrangements is very large, then you would want to jump to the $r^{th}$ configuration, without iterating through all $[0, 1, ..., r]$ configurations. This is coded into the approach to minimize iterating. 

Also, since the number of possible configurations increases with increased complexity of the binary tree, we give increased probability of selection to trees with more complexity, in order to enforce the quality that all equations be equally likely. This is implemented by changing the probability weights when randomly selecting the value for $i$. Since there are far more complex equations than simple ones, SRURGS will typically generate equations close to its user-specified upper bound of  complexity.

\subsection{Avoiding redundant calculations}
After accounting for arithmetic simplifications and rearrangement, many equations are identical to many other equations within the search space. This issue was thoroughly examined by the authors of another SR algorithm, PGE, who discussed methods to remove the duplicate equations. Their setup is quite complex, deploying the use of an n-ary tree representation along with partial ordering of subtree building blocks in order to avoid needless duplicate computations. However, they do not show any data demonstrating that this approach actually results in improved performance. This problem was also considered by \cite{Cozad2018}, who provide data supporting the notion that certain redundancy eliminating constraints in the MINLP context can decrease the run-time of their algorithm to 20\% of its baseline run-time. For our purpose, we remove equations whose evaluation results in floating point error and we apply the function simplification method of the symbolic computing library, Sympy \cite{sympy}. We believe this to be a pragmatic approach that is good enough. In addition, equivalence checking of arithmetic equations is known to be an NP-hard problem \cite{Ghodrat:2005} so we do not want to fixate on it.

\subsection{Including fitting constants}
Fitting parameters have been found to significantly affect the performance of symbolic regression routines, increasing their ability to solve more complex problems \cite{Kommenda, Kommenda2}. The use of fitting constants will improve the fit of many equations, increasing the number of equations which fit the equation to within our tolerance for error. Using fitting parameters shifts the shape of the equation, adjusting the relative contribution of different portions of the tree. Fitting parameters are equivalent to other terminals in the sense that they only occupy leaves in the tree. There are different approaches to determining the appropriate value of fitting constants, and there has been work performed to determine the appropriate method in the symbolic regression context \cite{constant_optimization}. These workers found that the two most effective methods for constant optimization were the Levenburg-Marquardt algorithm and the Nelder-Mead method. In our code, we use the Levenburg-Marquardt algorithm. This algorithm is used both in pySRURGS, our python code, and also the SRGP package we use for comparison.

\subsection{Parallel computing}
Because each iteration of SRURGS is independent from the previous, it is easy to distribute the computations. Our project, pySRURGS, saves the results to a SQLite backed dictionary object, so parallel produce results can be joined with duplicate computations erased. For our experiments, we distribute the jobs across multiple computers, leveraging an SQL database to track our computing jobs. Worker computers would query the database, retrieve a pending job, run computations, then store the results for transfer. Output files were sent from each computer to a central data storage, where the results are analyzed and statistical analyses are generated. At times, we leveraged cloud computing infrastructures, and it was very straightforward for us to deploy pySRURGS on these environments with many cores.

\subsection{Generating benchmark experiments}
We randomly generate 100 benchmark problems. The randomly generated benchmark problems are found online as a supplemental material to this paper. For all the problems, we specify that $N=200$. For these twenty problems, the permitted functions are \{+, \--, /, *\}. For the latter eighty problems, the permitted functions are \{+, \--, /, *, \^{}, exp, sin, sinh\}. Fitting parameters are randomly assigned values between [-10,10]. The dependent variables are sampled from a domain of [0, 10]. The maximum number of dependent variables we permit is 5. The maximum number of fitting parameters we permit is 5. For each problem, we sample 100 data points. When solving a symbolic regression problem, we use the same setup to guess the solution as was used in randomly generating the solution. Though this is not realistic because in reality we would not know which functions would be an appropriate basis for search, we apply this approach throughout so SRGP and SRURGS are on roughly equal footing. 

\subsection{Comparing genetic programming with uniform random global search}
We compare SRURGS with an implementation of SRGP using a heavily modified version of the DEAP symbolic regression tutorial script\footnote{\url{https://https://github.com/DEAP/deap}, the modified script is found in the pySRURGS github repository as the SRGP.py file.}. The SRGP algorithm is configured as follows. Individuals are spawned with a maximum height of 4. The maximum height of an expression mutation is 2. We use tournament selection with a tournament size of 5. The probability of crossover is 0.7 and the probability of mutation is 0.3. In genetic algorithms, two important tuning parameters are the size of the population and the number of generations the population is to evolve. The number of equations considered is not strictly ($population$ x $generations$), as some members of the parent population persist into the next generation. Therefore, we run the SRGP not for a set number of generations, but for as many generations as needed to produce our desired number of unique individuals. We run SRGP with a population of 100 for a variable number of generations till we create 1000 unique individuals. We use a modified version of the eaSimple algorithm from DEAP. When SRGP encounters an offspring that evaluates to a floating point error, it keeps the individual in the offspring, but it does not consider this individual as part of the 1000 unique individuals needed to complete its run. 

We use a paired t-test to compare SRGP against SRURGP. We take $p=0.05$ as our level of statistical significance. Since our main aim is to determine whether SRGP is superior to SRURGS, and we do not need to actually solve these benchmark problems, we can run our computations for fewer iterations that would be required to reach a highly accurate solution. We can then compare the performance of SRGP and SRURGS on this shorter computational task. We want to ensure that the algorithms being considered are on equal terms, therefore we share code between our pySRURGS package and our SRGP script, using the same method to simplify equations and store them in a persistent dictionary, which allows us to check whether an equation has been considered previously.

\begin{figure}[h]
    \centering
    \includegraphics[scale=1]{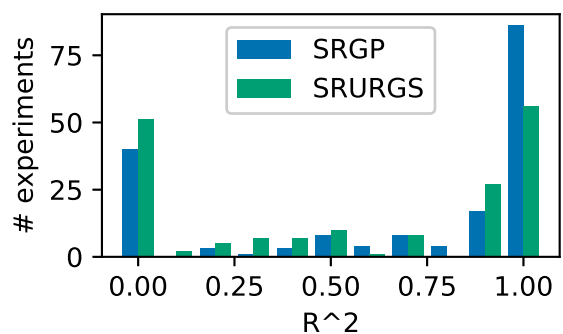}
        \caption{A histogram showing performance of SRURGS and SRGP on the first twenty benchmark problems}
        \label{fig:histo1}
\end{figure}

\begin{figure}[h]
    \centering
    \includegraphics[scale=1]{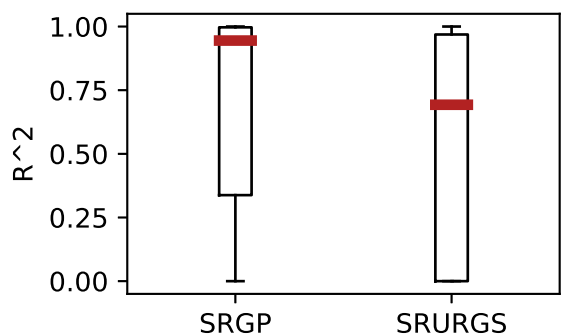}
    \caption{A boxplot comparing the $R^2$ of results for SRGP with those of SRURGS on the first twenty benchmark problems}
\label{fig:boxplot1}
\end{figure}

\section{Results and Discussion}

We will begin with a discussion of SRURGS' performance on a benchmark problem commonly considered in the symbolic regression literature, the quartic polynomial problem. We then discuss the results of our experiments against the randomly generated benchmark problems.

\subsection{The quartic polynomial toy problem}
To examine the nature of the search space within symbolic regression, we leverage the SRURGS code base in order to enumerate the solution space of the quartic polynomial problem, a commonly used test problem, and create a heatmap that highlight the performance of solutions as we iterate through the possible genotypes for a contiguous subset of the solution space. This is shown in Fig. \ref{fig:quartic_polynom_domain}. The index for the function configuration is $r$ and the index for the terminals configuration is $s$. No functions of arity one are permitted but (+, -, /, *, \^{}) are permitted as functions of arity two. Where $R^2$ evaluates to an invalid value, it has been set to zero. This plot corresponds to the fourth binary tree ($i=3$), where the domain of $r$ is 0 to 125 and the domain of $s$ is 0 to 81. We find that there is a certain periodicity to the solution space, but there are no clear trends that a search algorithm can use to traverse the terrain. Solutions are either identical to their neighbours, presumably because neighboring equations are equivalent after simplification or dramatically different. Most solutions are poor, though there are a fair number of good solutions. Note that we have intentionally devised an idealized search space, prohibiting functions of arity one and including a power function. We also allow only two fitting parameters. Fig. \ref{fig:quartic_polynom_result} shows how well SRURGS performed on the quartic polynomial problem.

\subsection{Performance on 100 benchmark problems}
We first consider the distribution of performance of the algorithms against the 100 problems. When considering the first twenty problems, which are simpler owing to the lack of functions of arity one, SRURGS performs slightly worse. The median and average $R^2$ for SRURGS over the 20 problems were 0.69 and 0.56 while those of SRGP were 0.94 and 0.68, respectively. We note that the average performances were quite poor for both algorithms. The p-value for the paired t-test was $5.1E-9$, demonstrating significance. The difference in distribution of performance is shown in Fig. \ref{fig:histo1}, where the most populated histogram bin in the SRGP distribution is that of the best performing models. By contrast, SRURGS split roughly 50-50 between the best and worst models. In Fig. \ref{fig:boxplot1}, we see that the interquartile range of SRGP results are better than those of SRURGS. Note that these results do not speak to the algorithm's ability to find better solutions when used with longer runtimes. We have done adhoc experiments, and SRURGS typically gives an $R^2$ approximately equal to 1 when run for 20000 iterations on problems of this complexity.

\begin{figure}[h]
    \centering
    \includegraphics[scale=1]{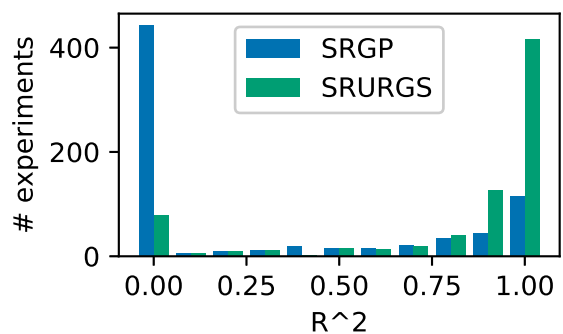}
        \caption{A histogram comparing the performance of SRURGS and SRGP on the latter eighty benchmark problems}
        \label{fig:histo2}
\end{figure}

\begin{figure}[h]
\centering
    \includegraphics[scale=1]{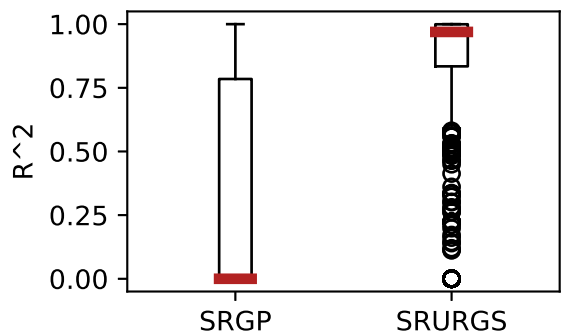}
    \caption{A boxplot comparing the $R^2$ of results for SRGP with those of SRURGS on the latter eighty benchmark problems}
\label{fig:boxplot2}
\end{figure}

The results for the latter eighty problems are similarly shown in Fig. \ref{fig:histo2} and Fig. \ref{fig:boxplot2}. These results show that with functions of arity one included, SRURGS is dramatically better than SRGP and performs even better than it did on the first twenty problems. The median and average $R^2$ for SRURGS over the 20 problems were 0.97 and 0.81 while those of SRGP were 0.00 and 0.31, respectively. The p-value for the paired t-test was $5.3E-124$, demonstrating significance. The difference in distribution of performance is shown in Fig. \ref{fig:histo2}, where the most populated histogram bin in the SRGP distribution is that of the worst performing models. By contrast, SRURGS generated mostly the best. In Fig. \ref{fig:boxplot2}, we see that the interquartile range of SRGP is broad and centered near zero, while SRURGS had a narrow interquartile range focused about 1. Since the performance of SRURGS is determined only by the proportion of solutions within the search space that are good, this plot is evidence that the search spaces for these seemingly more complex problems have a greater proportion of good solutions than do those of the more simple problems. These results also suggest that SRGP's evolutionary machinery is impeding its search of the solution space when certain functions of arity one are permitted.

\subsection{Limitations}
We wonder whether the comparison between the algorithms is truly fair. We are comparing algorithms using only the number of equations they consider in their search and pay no heed to the time they take in delivering their results. In addition, SRURGS is directly limited in the complexity of equations it is permitted to consider, while SRGP is not. Also, we have to choose parameters for SRGP in order to make the configuration of SRGP reasonable given our choice of parameters in generating the benchmark problems. These parameters include the complexity of the randomly generated initial members of the population, and the size of mutations when they happen. We have made every attempt to check that SRGP output is reasonable, but there is a degree of uncertainty associated with this hand tuning. It is computationally impractical to replicate our experiments for a large number of configuration permutations for SRGP, so we refrain from attempting a more broad set of experiments. It seems plausible that SRGP performs better than random search for problems with simpler solution spaces, but that is hardly praise, and it would be difficult to know when it is appropriate to limit the search space in this manner when approaching a symbolic regression problem.

\section{Conclusions}
We presented SRURGS, a novel methodology for symbolic regression. We demonstrated that SRURGS can be used to generate benchmark problems in symbolic regression. We expressed skepticism about the ubiquity of SRGP in SR problems despite the shadow of the NFLT. SRGP proponents may argue that a well tuned SRGP can outperform SRURGS, but to tune an SRGP would require significant computational effort that may be better spent running SRURGS. SRURGS tends to examine equations at the upper end of its complexity bounds. Practitioners can get around this issue by starting their search with simple equations then progressively increasing the permitted complexity of their search. We run 1000 experiments over 100 randomly generated benchmark problems comparing SRURGS with SRGP, demonstrating that SRURGS is a competitive alternative to SRGP.

\section{Compliance with Ethical Standards}
This article does not contain any studies with human participants or animals performed by any of the authors. Conflict of Interest: Sohrab Towfighi declares that he has no conflict of interest. The algorithm of Tychonievich was reproduced with permission from the author.

%
%

\bibliographystyle{spbasic}      
\bibliography{template}

\end{document}